\documentclass[accepted]{uai2023} 

\usepackage[american]{babel}

\usepackage{natbib} 
\usepackage{mathtools} 
\usepackage{booktabs} 
\usepackage{tikz} 
\usepackage{amsmath,amssymb}

\newtheorem{prop}{Proposition}
\newtheorem{defi}{Definition}
\newtheorem{rem}{Remark}

\newtheorem{cor}{Corollary}

\title{Robust Statistical Comparison of Random Variables \\with Locally Varying Scale of Measurement} 

\author[1]{\href{mailto:<christoph.jansen@stat.uni-muenchen.de>?Subject=Your UAI 2023 paper}{Christoph~Jansen}{}}
\author[1]{Georg~Schollmeyer}
\author[1]{Hannah~Blocher}
\author[1]{Julian~Rodemann}
\author[1]{Thomas~Augustin}
\affil[1]{%
    Department of Statistics\\
    Ludwig-Maximilians-Universität\\
    Munich, Bavaria, Germany
}
  
\begin{document}
\maketitle

\begin{abstract}
Spaces with locally varying scale of measurement, like multidimensional structures with differently scaled dimensions, are pretty common in statistics and machine learning. Nevertheless, it is still understood as an open question how to exploit the entire information encoded in them properly. We address this problem by considering an order based on (sets of) expectations of random variables mapping into such non-standard spaces. This order contains stochastic dominance and expectation order as extreme cases when no, or respectively perfect, cardinal structure is given.  We derive a (regularized) statistical test for our proposed generalized stochastic dominance (GSD) order, operationalize it by linear optimization, and robustify it by imprecise probability models. Our findings are illustrated with data from multidimensional poverty measurement, finance, and medicine.  
\end{abstract}

\section{Introduction}\label{sec:intro}
Numerous challenges in statistics and machine learning can -- at least theoretically -- be broken down to comparing random variables $X,Y: \Omega \to A$ mapping between measurable spaces  $(\Omega, \mathcal{S}_1)$ and $(A,\mathcal{S}_2)$. Consequently, much attention has been paid to find and apply well-founded \textit{stochastic orderings} enabling such comparison. Examples range from classifier comparisons (e.g.,~\cite{d2006}, \cite{ccbdm2017}, or~\cite{bsjn2023}) over ranking risky assets (e.g., \cite{c2015}) to deriving optimal (generalized) Neyman-Pearson tests (e.g., \cite[§7.4]{Augustin:Walter:Coolen:2014:itip:stat}). 

In the traditional case where the context allows to specify both a probability $\pi$ on $\mathcal{S}_1$, and a \textit{cardinal} scale $u: A \to \mathbb{R}$ representing the structure on $A$, a common order $\succsim_{E(u)}$ on $
    \bigr\{X \in A^{\Omega}: u \circ X \in  \mathcal{L}^1(\Omega ,\mathcal{S}_1, \pi)\bigl\}$
is obtained by setting $(X,Y) \in \succsim_{E(u)}$ if and only if
\begin{equation} \label{EU}
    \mathbb{E}_{\pi}(u \circ X)=\int_{\Omega}u \circ X d\pi\geq \int_{\Omega}u \circ Y d\pi= \mathbb{E}_{\pi}(u \circ Y).
\end{equation}
Here, random variables are ranked according to the expectations of their numerical equivalents induced by the scale $u$. We take the following perspective: This order $\succsim_{E(u)}$ would be the desired order if we were confronted with a problem under pure \textit{aleatoric} uncertainty where an (objective) probability measure $\pi$ and a cardinal scale $u$ \textit{were} available.\footnote{The term "aleatoric uncertainty" seems adequate only when $\pi$ refers to a stochastic phenomenon. However, $\pi$ might as well represent subjective beliefs which can be formalized by a probability measure such as, e.g., in the Bayesian school of thought.}

Our paper addresses all situations where, in addition, \textit{epistemic uncertainty} has to be taken into account. Then, such single $\pi$ and $u$ (and consequently the expectations in~(\ref{EU})) are not available, rendering a comparison by $\succsim_{E(u)}$ impossible. This non-availability corresponds to two facets (e.g.~\cite{hullermeier2021aleatoric}) of epistemic uncertainty: Referring to $\pi$, \textit{approximation} uncertainty arises since -- as common in statistics --  only samples of the considered variables are available.\footnote{In Section~\ref{robtest} we go beyond approximation uncertainty and consider robustification by a candidate set of probabilities.} Concerning $u$, on the other hand, \textit{model} uncertainty is assumed to occur from weakly structured order information, making a non-singleton 
\textit{set}~$\mathcal{U}$ of candidate scales compatible with the structure on $A$.

Naturally, such situations can be approached in two steps: Focusing-- in the first step -- on model uncertainty, and thus assuming $\pi$ still to be known, the order $\succsim_{E(u)}$ can be weakened to a \textit{preorder} $\succsim_{(\mathcal{U},\pi)}$ on
\begin{equation*}
   \Bigl\{X \in A^{\Omega}: u \circ X \in  \mathcal{L}^1(\Omega ,\mathcal{S}_1, \pi)~\forall u \in \mathcal{U}\Bigr\}
\end{equation*}
by setting $(X,Y) \in \succsim_{(\mathcal{U},\pi)}$ if and only if Inequality~(\ref{EU}) holds for all candidate scales $u \in \mathcal{U}$. Depending on the concrete choice of the set $\mathcal{U}$, the relation $\succsim_{(\mathcal{U},\pi)}$ has some prominent special cases: If $A$ is equipped with a preorder, and $\mathcal{U}$ is the set of all functions that are bounded and isotone w.r.t.~this preorder, then $\succsim_{(\mathcal{U},\pi)}$ is (essentially) equivalent to (first-order) stochastic dominance. In contrast, if  $(A ,\mathcal{S}_2)=(\mathbb{R}, \mathcal{B}_{\mathbb{R}})$ and $\mathcal{U}$ consists of all bounded and \textit{concave} functions, then $\succsim_{(\mathcal{U},\pi)}$ (essentially) corresponds to second-order stochastic dominance. 

If -- in a second step -- information about $\pi$ comes only from samples from the distributions of $X$ and $Y$, then, instead of the order $\succsim_{(\mathcal{U},\pi)}$, one has to rely on the corresponding empirical version. Then, a statistical test is needed to control the probability of wrong conclusions from the data. 

\textbf{Motivation of our work:} The main goal of the present work is to provide scientists from different fields of application with an inference methodology for the robust analysis of systematic distributional differences within a population. On the one hand, it is important to go beyond a simple comparison of location measures, similar to the case of classical stochastic dominance. On the other hand, we want to take into account the fact that classical (first-order) stochastic dominance systematically ignores potentially available metric information. We achieve this by a generalized stochastic dominance ordering (GSD), which is based on the flexible concept of preference systems. Specifically, we propose a nonparametric permutation test for subgroup comparison that robustifies (therefore further weakening the already parsimonious assumptions) towards the often-criticized assumption of exactly representative sampling.

\textbf{Our contribution}: We consider generalized stochastic dominance (GSD) that ensures exploiting the entire information encoded in data with locally varying scale of measurement. For that purpose,  we (primarily) focus, technically speaking, on that specific class of preorders $\succsim_{(\mathcal{U},\pi)}$ where $\mathcal{U}$ is the set of representations of a \textit{preference system} (cf.~Sections~\ref{intro} to~\ref{gsd}).  Then, using linear optimization, we derive a corresponding (regularized) test (cf.~Section~\ref{testdom}) and robustify it relying on imprecise probabilities  (cf.~Section~\ref{robtest}).  Particularly, our framework allows handling multidimensional structures with differently scaled dimensions in an information-efficient way (cf.~Section~\ref{multi}). We illustrate this with data from multidimensional poverty measurement, finance, and medicine (cf.~Section~\ref{application} and Supp.~D) and conclude with a brief discussion (cf.~Section~\ref{conrem}). The proofs of Propositions~\ref{obs1} to \ref{simplification}, and Corollary~\ref{cor1} can be found in the supplementary material (cf.,~Supp.~A). Our code is available under: \url{https://github.com/hannahblo/Robust_GSD_Tests}

\textbf{Related work:} Work on tests and/or checking algorithms for stochastic dominance (SD) outside preference systems includes \cite{f1989,Mosler1991,m1995,bd2003,sja2017,ro2019,chetverikov_wilhelm_kim_2021}. Optimization under SD constraints was recently considered by, e.g., \cite{Dai:StochDomConst:AIStats:2023}.  
Preference systems and related structures are discussed in a decision theoretic context in \cite{p2013} and \cite{jsa2018,jbas2022}. A test for GSD in the special case of a preference system arising from multiple quality metrics in classifier comparison is discussed in~\cite{jnsa2022}. 

Neighborhood models that are used to robustify tests are studied in e.g., \cite{Destercke:Montes:Miranda:Dist:Compar:2022,Augustin:Schollmeyer:StatSc:2021,Montes:Miranda:Destercke:2020:PartII}. Among others, \cite{Maua:deCampos:2021,Cabanas:Antonucci:et:al:CREDICI:2020,Maua:Cozman:2020} study credal networks as robustifications of Bayesian networks, and, e.g., \cite{Utkin:Konstantinov:RF:Cont:NeuNet:2022,ra2022,CarranzaAlarcon:Destercke:PattRec:2021,Utkin:Deep:Forest:ExpSsAppl:2020,Abellan:Mantas:Castellano:Moral-Garcia:ExpSsAppl:2018} have proposed robustifications and extensions of other machine learning procedures like forests or discriminant analyses by imprecise probabilities.

Accounting for both approximation uncertainty and model uncertainty is in line with recent deliberations in uncertainty quantification (e.g.,~\cite{malinin2018predictive, hullermeier2021aleatoric, bengs2022pitfalls,Huellermeier:Destercke:Shaker:UAI:2022}). 

\section{Background \& Preliminaries} \label{intro}
We will consider \textit{binary relations} at several points, relying on the following notation and terminology: A binary relation $R$ on a set $M \neq \emptyset$ is a subset of the Cartesian product of $M$ with itself, i.e.~$R \subseteq M \times M$. $R $ is called \textit{reflexive}, if $(a,a) \in R$, \textit{transitive}, if $(a,b),(b,c) \in R\Rightarrow(a,c) \in R$, \textit{antisymmetric}, if $(a,b),(b,a) \in R\Rightarrow a=b$, \textit{complete}, if $(a,b) \in R$ or $(b,a) \in R$ (or both) for arbitrary elements $a,b,c \in M$. A \textit{preference relation} is a binary relation that is complete and transitive; a \textit{preorder} is a binary relation that is reflexive and transitive; a \textit{linear order} is a preference relation that is antisymmetric; a \textit{partial order} is a preorder that is antisymmetric. If $R$ is a preorder, we denote by $P_R \subseteq M \times M$ its \textit{strict part} and by $I_R \subseteq M \times M$ its \textit{indifference part}, defined by $ (a , b) \in P_{R} \Leftrightarrow (a , b) \in R \wedge (b , a) \notin R,$
and $(a , b) \in I_{R} \Leftrightarrow (a , b) \in R \wedge (b , a) \in R$.

This leads us to the central ordering structure under consideration in the present paper, namely \textit{preference systems}. These formalize the idea of spaces with locally varying scale of measurement and were introduced in~\cite{jsa2018}.\footnote{For a study on representation results of the related concept of \textit{incomplete difference preorders} see, e.g.,~\cite{p2013}.}
\begin{defi}\label{ps}
Let $A \neq \emptyset$ be a set, $R_1 \subseteq A \times A$ a preorder on $A$, and
 $R_2 \subseteq R_1 \times R_1$ a preorder on $R_1$. The triplet $\mathcal{A}=[A, R_1 , R_2]$ is then called a \textbf{preference system} on $A$. We call $\mathcal{A}$ \textbf{bounded}, if there exist $a_*,a^* \in A$ such that $(a^*,a) \in R_1$, and $(a,a_*) \in R_1$ for all $a \in A$, and $(a^*,a_*) \in P_{R_1}$. Moreover, the preference system $\mathcal{A}'=[A', R_1' , R_2']$ is called \textbf{subsystem} of $\mathcal{A}$ if $A' \subseteq A$, $R_1'\subseteq R_1$, and $R_2'\subseteq R_2$. In this case, we call $\mathcal{A}$ a \textbf{supersystem} of $\mathcal{A}'$.
\end{defi}
The concrete definition of a preference system now also makes it possible to concretize the idea of a space with \textit{locally varying scale of measurement}: While the relation $R_1$ formalizes the available ordinal information, i.e. information about the arrangement of the elements of $A$, the relation $R_2$ describes the cardinal part of the information in the sense that pairs standing in relation are ordered with respect to the intensity of the relation. Thus, intuitively speaking, the set $A$ is locally almost cardinally ordered on subsets where $R_1$ and $R_2$ are very dense, while on subsets where $R_2$ is sparse or even empty, locally at most an ordinal scale of measurement can be assumed. A natural example is multi-dimensional structures with differently scaled dimensions, such as those that appear in the poverty analysis application discussed in Section~\ref{application}: While variables like education can be assumed to have only ordinal scale of measurement, a variable like income is rather metrically scaled.

To ensure that $R_1$ and $R_2$ are compatible, we use a consistency criterion for preference systems relying on the idea that both relations should be simultaneously representable.
\begin{defi} \label{consistency}
The preference system $\mathcal{A}=[A, R_1 , R_2]$ is \textbf{consistent} if there exists a \textbf{representation} $u:A \to \mathbb{R}$ such that for all $a,b,c,d \in A$ we have:
\begin{itemize}\itemsep1.5mm
\item[i)] If we have that $ (a , b) \in R_1$, then it holds that $u(a) \geq u(b)$, where equality holds if and only if $(a,b)\in I_{R_1}$.
\item[ii)] If we have that $((a , b),(c,d)) \in R_2$, then it holds that $u(a) -u(b) \geq u(c)-u(d)$, where equality holds if and only if $((a,b),(c,d))\in I_{R_2}$.
\end{itemize}
The set of all representations of $\mathcal{A}$ is denoted by $\mathcal{U}_{\mathcal{A}}$. 
\label{consistent}
\end{defi}
Especially when regularizing our test statistic in Section~\ref{testdom}, normalized versions of the set $\mathcal{U}_{\mathcal{A}}$ play a crucial role. 
\begin{defi} \label{granularity}
Let $\mathcal{A}=[A, R_1 , R_2]$ be a consistent and bounded preference system with $a_*,a^*$ as before. Then 
$$\mathcal{N}_{\mathcal{A}}:= \Bigl\{u  \in \mathcal{U_{\mathcal{A}}}: u(a_*)=0 ~\wedge ~ u(a^*)=1 \Bigl\}$$
is called the \textbf{normalized representation set} of $\mathcal{A}$. Further, for $\delta \in [0,1)$, we denote by $\mathcal{N}^{\delta}_{\mathcal{A}}$ the set of all $u \in \mathcal{N}_{\mathcal{A}}$ with
 $$ u(a)-u(b) \geq \delta ~~~\wedge ~~~ u(c)-u(d) -u(e) + u(f) \geq \delta $$ 
 for all $(a,b) \in P_{R_1}$ and for all $((c,d),(e,f)) \in P_{R_2} $. We call $\mathcal{A}$ \textbf{$\delta$-consistent} if $\mathcal{N}^{\delta}_{\mathcal{A}} \neq \emptyset$.
\end{defi}
We conclude the section with an immediate observation of the connection between consistency and $0$-consistency.
\begin{prop}\label{obs1}
Let $\mathcal{A}=[A, R_1 , R_2]$ be a bounded preference system. Then $\mathcal{A}$ is consistent if and only if it is $0$-consistent.
\end{prop}
\section{Regularization} \label{regu}
\begin{figure}[ht]
\centering
\includegraphics[width=7cm]{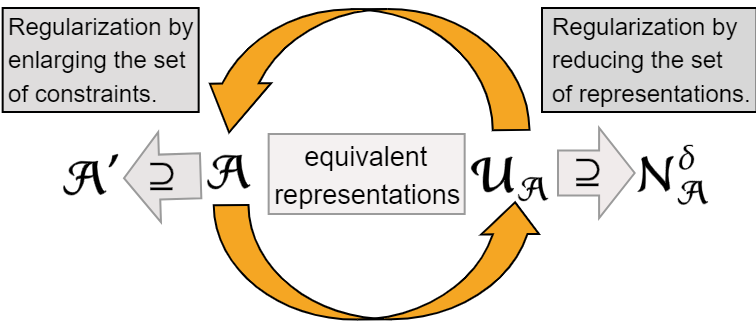}
\caption{Two ways for regularizing a preference system.}
\label{fig1}
\end{figure}

We now discuss some thoughts on regularization in preference systems. Since our later considerations primarily concern statistical testing, regularization then aims at making the test statistic more sensitive, i.e., to increase discriminative power. In contrast to the usually advocated Thikonov-type regularization, here we think in terms of Ivanov-type regularization that constraints the space of functions (in our case $\mathcal{N}_\mathcal{A}$) over which later our optimization is done (cf., Section~\ref{teststatistic} where our test statistic is introduced as an infimum type test statistic). Beyond the different more or less equivalent ways of representing regularization in a Thikonov-, Ivanov- or in a Morozov-type style (cf. \cite{ora2016}), here additionally, two different ways of implementing regularization are conceivable: On the one hand, an \textit{order-theoretic} regularization could be carried out by extending the considered preference system by additional comparable pairs (or pairs of pairs) to a consistent super system. On the other hand, a \textit{parameter-driven} regularization could be performed to reduce the set of representations of the preference system. Both ways are schematically compared in Figure~\ref{fig1}.

Both approaches have their own strengths and weaknesses: In the case of order-theoretic regularization, the influence of the regularization on the content-related question can be controlled very precisely. However, this comes at the price that the concrete mathematical influence of the regularization can only be characterized with difficulty. The situation tends to be reversed in the case of parameter-driven regularization: Here, it is straightforward -- by choosing larger and larger parameter values -- to control the mathematical influence of the regularization. However, an interpretation of the regularization in the context of the content-related question is less direct than in the first case. Nevertheless, a possible interpretation in a decision-theoretic context is given in~\cite{jsa2018} by establishing a connection to Luce's \textit{just noticeable differences} \citep{luce}. In this paper, we focus on parameter-driven regularization since, for regularization of the test statistic used later, the interpretation of the parameter is of secondary importance.
\section{Generalized Dominance} \label{gsd}
As indicated at the outset, we now turn to a stochastic order between random variables with values in a preference system. This order rigorously generalizes stochastic dominance in the sense that it optimally exploits also the partial cardinal information encoded in these spaces.  Therefore, it is neither limited to a purely ordinal analysis as first-order stochastic dominance nor requires perfect cardinal information as second-order stochastic dominance. Consequently, in cases without any cardinal information, i.e., where $R_2$ is the trivial preorder, the considered order reduces back to the first-order stochastic dominance.

We start by introducing some additional notation: For $\pi$ a probability measure on $(\Omega ,\mathcal{S}_1)$ and $\mathcal{A}$ a consistent preference system, we define by $\mathcal{F}_{(\mathcal{A},\pi)}$ the set
\begin{equation*}
    \Bigr\{X \in A^{\Omega}: u \circ X \in  \mathcal{L}^1(\Omega ,\mathcal{S}_1, \pi)~\forall u \in \mathcal{U}_{\mathcal{A}}\Bigl\}.
\end{equation*}
We then can define the following preorder on $\mathcal{F}_{(\mathcal{A},\pi)}$.
\begin{defi}\label{GSD}
Let $\mathcal{A}=[A, R_1 , R_2]$ be consistent. For $X,Y \in \mathcal{F}_{(\mathcal{A},\pi)}$, we say  $Y$ is \textbf{$(\mathcal{A},\pi)$-dominated} by $X$ if 
$$\mathbb{E}_{\pi}(u \circ X) \geq \mathbb{E}_{\pi}(u \circ Y)$$
 for all $u \in \mathcal{U}_{\mathcal{A}}$. The induced relation is denoted by $R_{(\mathcal{A},\pi)}$ and called \textbf{generalized stochastic dominance (GSD)}.
\end{defi}
We have the following immediate simplification if the underlying preference system $\mathcal{A}$ is additionally bounded.
\begin{prop} \label{normalized}
If $\mathcal{A}$ is consistent and bounded with $a_*,a^*$ as before, then $(X,Y) \in R_{(\mathcal{A},\pi)}$ iff
\begin{equation}
\forall u \in \mathcal{N}_{\mathcal{A}}: \mathbb{E}_{\pi}(u \circ X) \geq \mathbb{E}_{\pi}(u \circ Y).
\end{equation}
\end{prop}
\section{Testing for Dominance} \label{testdom}
 Throughout this section, let $\mathcal{A}=[A, R_1 , R_2]$ be \textit{consistent} and \textit{bounded} with $a_*,a^* \in A$ as in Definition~\ref{ps}. 

We now turn to the statistical version of our investigation, where we do not know the underlying probability $\pi$ but \textit{i.i.d.} samples $\mathbf{X}=(X_1, \dots , X_n)$ and $\mathbf{Y}=(Y_1, \dots , Y_m)$ of $X$ and $Y$ are available. The fundamental question now is when we can, with a certain error probability, conclude from this information that $X,Y \in \mathcal{F}_{(\mathcal{A},\pi)}$ are in relation with respect to the GSD-relation $R_{(\mathcal{A},\pi)}$. Constructing a corresponding test, we first need to be clear about appropriate statistical hypotheses. Ideally, we would be interested in the following pair of hypotheses:
\begin{equation}
    H_0^{id}: (X,Y) \notin R_{(\mathcal{A},\pi)}~~~\text{\textbf{vs.}}~~~ H_1^{id}: (X,Y) \in R_{(\mathcal{A},\pi)} 
\end{equation}
In the pair $(H_0^{id}, H_1^{id})$ of hypotheses -- as intended in a statistical test -- the question actually of interest would be formulated as the alternative hypothesis. Then, the probability of falsely assuming it to be true could be controlled by the significance level. Unfortunately, similar to the situation of classical stochastic dominance as described, e.g., in \cite{bd2003} and further investigated in~\cite{shaked2007stochastic}, or generally in the context of bio-equivalence testing (e.g., \cite{Brown:Hwang:Munk:BioEq:AnnStat:1997}), the hypothesis $H_0^{id}$ seems to be too broad for a meaningful analysis, in the sense that the most conservative scenario under $H_0^{id}$ is not clearly specifiable.\footnote{The problem is due to the fact that the relation $R_{(\mathcal{A},\pi)}$ is a \textit{partial} order. Compare also~\cite[p.~24-25]{sja2017}.} For this reason, we choose a pair of alternatives that deviates slightly from the actual question of interest and afterwards try to make the deviation from the actual pair of hypotheses of interest assessable by testing with the variables in reversed roles. The modified pair of hypotheses looks as follows:
\begin{equation}
    H_0: (Y,X) \in R_{(\mathcal{A},\pi)}~~~\text{\textbf{vs.}}~~~ H_1: (Y,X) \notin R_{(\mathcal{A},\pi)} \label{praghyp}
\end{equation}
The advantage of the pair $(H_0,H_1)$ is that a worst-case analysis of the distribution of a suitable test statistic under $H_0$ is possible: The test statistic would have to be analyzed under the most conservative case within $H_0$, namely $\pi_X = \pi_Y$, with $\pi_X$ and $\pi_Y$ the image measures of $X$ and $Y$ under $\pi$. The drawback to the pair $(H_0,H_1)$ is that in the case of rejection of $H_0$ we can only control the erroneous conclusion on $(Y,X) \notin R_{(\mathcal{A},\pi)}$ (and not the one actually of interest on $(X,Y) \in R_{(\mathcal{A},\pi)}$) in its probability by the significance level. To mitigate this effect, we can test with the pair $(H_0,H_1)$ of hypotheses additionally with $X$ and $Y$ in reversed roles. 
\subsection{The Choice of the Test Statistic}\label{teststatistic}
For defining an adequate test statistic, we first note that -- due to the boundedness of $\mathcal{A}$ and Proposition~\ref{normalized} -- it holds $(X,Y) \in R_{(\mathcal{A},\pi)}$ if and only if
\begin{equation}\label{theoretical}
D(X,Y):= \inf_{u\in \mathcal{N}_{\mathcal{A}}}\left(\mathbb{E}_{\pi}(u \circ X) - \mathbb{E}_{\pi}(u \circ Y) \right)\geq 0,
\end{equation}
i.e., if the infimal expectation difference with respect to the available information is at least zero. Thus, a straightforward test statistic is the empirical version of $D(X,Y)$, i.e.,
\begin{equation*}
    d_{\mathbf{X},\mathbf{Y}}:\Omega \to \mathbb{R}
\end{equation*}
\begin{equation*}
    \omega \mapsto \inf_{u\in \mathcal{N}_{\mathcal{A}_{\omega}}}\sum_{z \in (\mathbf{X} \mathbf{Y})_{\omega}} u(z)\cdot(\hat{\pi}^{\omega}_X(\{z\})-\hat{\pi}^{\omega}_Y(\{z\}))
\end{equation*}
with, for $\omega \in \Omega$ fixed, $\hat{\pi}^{\omega}_X(\cdot):=\tfrac{1}{n}|\{i:X_i(\omega)\in \cdot\}|$ and $\hat{\pi}^{\omega}_Y(\cdot):=\tfrac{1}{m}|\{i:Y_i(\omega)\in \cdot\}|$ the observed empirical image measures of $X$ and $Y$,
$$(\mathbf{X} \mathbf{Y})_{\omega}= \{X_i(\omega):i\leq n\}\cup \{Y_i(\omega):i \leq m\} \cup \{a_*,a^*\},$$  
and $\mathcal{A}_{\omega}$ the subsystem of $\mathcal{A}$ restricted  to $(\mathbf{X} \mathbf{Y})_{\omega}$. If $d_{\mathbf{X},\mathbf{Y}}(\omega_0) \geq 0$ holds for some $\omega_0 \in \Omega$, we say there is \textit{in-sample GSD} of $X$ over $Y$ in the sample induced by $\omega_0$. {If the underlying space $\mathcal{A}$ is not too complex\footnote{A concrete sufficient condition  for consistency of $d_{\mathbf{X},\mathbf{Y}}$ is a finite VC dimension of the class of all indicator functions of the form $\{a \mid u(a) \geq c\}$ with $ u \in \mathcal{N}_\mathcal{A}$. This property is usually given, for example if we have finitely many dimensions which have itself a finite VC dimension. Therefore, especially in our applications of Section~\ref{application} consistency is guaranteed.} (under \textit{i.i.d.} within every subgroup) this test statistic converges to the true value of $D(X,Y)$ and is therefore an adequate test statistic for our test.

As a further test statistic, we consider a regularized version $d^{\varepsilon}_{\mathbf{X},\mathbf{Y}}$ of $d_{\mathbf{X},\mathbf{Y}}$: The infimum in the definition of $d_{\mathbf{X},\mathbf{Y}}$ is now only computed among $[0,1]$-normalized representations of $\mathcal{A}_{\omega}$ that distinguish between strictly related alternatives over some prespecified threshold value. In this way, the regularized test statistic is also sensitive for distinguishing situations under dominance regarding their \textit{extent} of dominance: While in-sample GSD (essentially) implies $d_{\mathbf{X},\mathbf{Y}}(\omega_0)=0$, it often holds $d^{\varepsilon}_{\mathbf{X},\mathbf{Y}}(\omega_0)>0$. Thus, for $V,W$ with $(\mathbf{V} \mathbf{W})_{\omega_0}=(\mathbf{X} \mathbf{Y})_{\omega_0}$ it might be that $d_{\mathbf{V},\mathbf{W}}(\omega_0)=0$ and  $d^{\varepsilon}_{\mathbf{X},\mathbf{Y}}(\omega_0)>d^{\varepsilon}_{\mathbf{V},\mathbf{W}}(\omega_0)>0$ and, hence, that under regularization $X$ (empirically) dominates $Y$ more strongly than $V$ dominates $W$.\footnote{As an example, in the situation of a preference system guaranteeing a totally ordered space (i.e., $R_2$ is the trivial preorder, $R_1$ is complete) where the laws of the random variables build a location family $\{ f(\cdot +c) \mid c \in \mathbb{R}\}$, the regularized statistic (with appropriately chosen $\delta $) will capture the  difference $\Delta = c-\tilde{c}$ between two populations distributed according to $f(\cdot + c)$ and $f(\cdot+ \tilde{c})$, respectively, whereas the non-regularized test will not.} 

Formally,  the regularized test statistic looks as follows:
\begin{equation*}
  \displaystyle  d^{\varepsilon}_{\mathbf{X},\mathbf{Y}}:\Omega \to \mathbb{R}
\end{equation*}
\begin{equation*}
\displaystyle    \omega \mapsto \inf_{u\in \mathcal{N}^{\delta_{\varepsilon}(\omega)}_{\mathcal{A}_{\omega}}}\sum_{z \in (\mathbf{X} \mathbf{Y})_{\omega}} u(z)\cdot(\hat{\pi}^{\omega}_X(\{z\})-\hat{\pi}^{\omega}_Y(\{z\}))
\end{equation*}
with $\varepsilon \in [0,1]$ and $
    \delta_{\varepsilon}(\omega):= \varepsilon \cdot\sup \{\xi:\mathcal{N}^{\xi}_{\mathcal{A}_{\omega}} \neq \emptyset\}.$ 
 Observe that $d_{\mathbf{X},\mathbf{Y}} = d^{0}_{\mathbf{X},\mathbf{Y}}$, i.e.,~the unregularized test statistic equals the regularized one if $\varepsilon=0$.
\subsection{A Permutation-Based Test}\label{sec:permutation_test}
As the distribution of $d_{\mathbf{X},\mathbf{Y}}$ and $ d^{\varepsilon}_{\mathbf{X},\mathbf{Y}}$ can not be straightforwardly analyzed, we utilize that under the above i.i.d.-assumption a permutation-based test (see, e.g.,~\cite{pg2012}) can be performed. For this, we assume we made observations of the i.i.d.~variables, i.e.,~we observed
\begin{eqnarray}\label{sample1}
    \mathbf{x}&:=&(x_1, \dots , x_n):=(X_1(\omega_0), \dots , X_n(\omega_0))
\\
\label{sample2}
    \mathbf{y}&:=&(y_1, \dots , y_m):=(Y_1(\omega_0), \dots , Y_m(\omega_0))
\end{eqnarray}
for some $\omega_0 \in \Omega$. The resampling scheme for analyzing the distributions of $d_{\mathbf{X},\mathbf{Y}}$ and $ d^{\varepsilon}_{\mathbf{X},\mathbf{Y}}$, respectively, can then be described by the following steps:

\textbf{Step 1:} Take the pooled data sample: $$\mathbf{w}:=(w_1, \dots , w_{n+m}):=(x_1, \dots , x_n, y_1 , \dots , y_m)$$ 
\textbf{Step 2:} Take all $k:=\binom{n+m}{n}$ index sets $I \subseteq \{1, \dots , n+m \}$ of size $n$. Evaluate $d_{\mathbf{X},\mathbf{Y}}$ resp. $d^{\varepsilon}_{\mathbf{X},\mathbf{Y}}$ for  $(w_i)_{i \in I}$   and  $(w_i)_{i \in \{1, \dots , n+m \} \setminus I}$ instead of  $\mathbf{x}$ and $\mathbf{y}$. Denote the evaluations by $d_I$ resp. $d_I^{\varepsilon}$.

\textbf{Step 3:} Sort all $d_I$ resp. $d_I^{\varepsilon}$ in increasing order to get $d_{(1)},\dots, d_{(k)}$ resp.~$d_{(1)}^{\varepsilon},\dots, d_{(k)}^{\varepsilon}$.

\textbf{Step 4:} Reject $H_0$ if $d_{\mathbf{X},\mathbf{Y}}(\omega_0)$ resp. $d^{\varepsilon}_{\mathbf{X},\mathbf{Y}}(\omega_0)$ is greater than $d_{(\ell)}$ resp. $d_{(\ell)}^{\varepsilon}
$, with $\ell := \lceil (1-\alpha) \cdot k\rceil$ and $\alpha$ the
significance level.

Note that, for large $\binom{n+m}{n}$, we can approximate the above resampling scheme by computing $d_I$ resp. $d_I^{\varepsilon}$ only for a large number $N$ of randomly drawn $I$. Moreover, note that only the \textit{i.i.d.} assumption is needed for the above test to be valid. (Precisely, it would already be enough to assume \textit{exchangeable} observations of both variables.)
\subsection{Computation of $\mathbf{d_{\mathbf{X},\mathbf{Y}}}$ and $\mathbf{d^{\varepsilon}_{\mathbf{X},\mathbf{Y}}}$}\label{sec:compu_test_statistic}
We show how the test statistics $d_{\mathbf{X},\mathbf{Y}}$ and $ d^{\varepsilon}_{\mathbf{X},\mathbf{Y}}$ can be computed in concrete cases. For that, we
consider samples $\mathbf{x}$ and $\mathbf{y}$ of the form ~(\ref{sample1}) and~(\ref{sample2}), and we assume w.l.o.g. that 
\begin{equation*}
    (\mathbf{X} \mathbf{Y})_{\omega_0}=\{z_1=a_*,z_2=a^*,z_3, \dots, z_s\}
\end{equation*}
Further, we denote by $C(\mathbf{x},\mathbf{y})$ the set of all vectors $(v_1, \dots , v_s,\xi) \in [0,1]^{s+1}$ such that $v_{1}=0$ and $v_{2}=1$ and for which it holds that
\begin{itemize}
    \item $v_i=v_j $ if $(z_i,z_j) \in I_{R_1}$,
    \item $v_i-v_j \geq \xi$ if $(z_i,z_j) \in P_{R_1}$,
    \item $v_k-v_l= v_r -v_t $ if $((z_k,z_l),(z_r,z_t)) \in I_{R_2}$
and
\item $v_k-v_l-v_r+v_t \geq \xi$ if $((z_k,z_l),(z_r,z_t)) \in P_{R_2}$.
\end{itemize}
Moreover, for $\xi_0 \in [0,1]$ fixed, we define $C_{\xi_0}(\mathbf{x},\mathbf{y})$ as $\{(v_1, \dots , v_s) \in [0,1]^s:(v_1, \dots , v_s, \xi_0) \in C(\mathbf{x},\mathbf{y})\}$, i.e., the set of all sample weights that respect the observed preference system and distinguish the strict part of its relations above a threshold of $\xi_0$. Both $C(\mathbf{x},\mathbf{y})$ and $C_{\xi_0}(\mathbf{x},\mathbf{y})$ are described by finitely many linear inequalities on $(v_1, \dots , v_s,\xi)$ resp. $(v_1, \dots , v_s)$. This allows to formulate Propositions~\ref{prop:xi_computation} and~\ref{prop:test_statitic_reg}. The first one demonstrates how to compute the maximum regularization threshold, whereas the second one captures the computation of $d_{\mathbf{X},\mathbf{Y}}$ and $ d^{\varepsilon}_{\mathbf{X},\mathbf{Y}}$.
\begin{prop}\label{prop:xi_computation}
For samples $\mathbf{x}$ and $\mathbf{y}$ of the form ~(\ref{sample1}) and~(\ref{sample2}) and $\varepsilon \in [0,1]$, we consider the linear program (LP)
\begin{equation}\label{maxdel}
    \xi \longrightarrow \max_{(v_1 , \dots , v_s,\xi)}
\end{equation}
with constraints $(v_1 , \dots , v_s,\xi) \in C(\mathbf{x},\mathbf{y})$. Denote by $\xi^*$ its optimal value. It then holds $\delta_{\varepsilon}(\omega_0)=\varepsilon \cdot \xi^*$.
\end{prop}
\begin{prop}\label{prop:test_statitic_reg}
For samples $\mathbf{x}$ and $\mathbf{y}$ of the form ~(\ref{sample1}) and~(\ref{sample2}) and $\varepsilon \in [0,1]$, we consider the following LP
\begin{equation}
    \sum_{\ell =1}^{s} v_{\ell}\cdot\Bigl(\tfrac{|\{i:x_i=z_{\ell}\}|}{n}-\tfrac{|\{i:y_i=z_{\ell}\}|}{m}\Bigr) \longrightarrow \min_{(v_1 , \dots , v_s)}
\end{equation}
with $(v_1 , \dots , v_s) \in C_{\varepsilon  \xi^*}(\mathbf{x},\mathbf{y})$, where $\xi^*$ is the optimal value of~(\ref{maxdel}).  Denote by opt$_{\varepsilon}(\mathbf{x},\mathbf{y})$ its optimal value. Then:
\begin{itemize}
    \item[i)] opt$_{\varepsilon}(\mathbf{x},\mathbf{y})=d^{\varepsilon}_{\mathbf{X},\mathbf{Y}}(\omega_0)$. 
    \item[ii)] It holds in-sample GSD of $X$ over $Y$ iff opt$_0(\mathbf{x},\mathbf{y})\geq 0$.
\end{itemize}
\end{prop}
\section{Robustified Testing using IP} \label{robtest}
Our test for GSD relies on i.i.d.~samples of the populations of actual interest. It thus can be based directly on the observed empirical distributions. We now show how \textit{imprecise probabilities (IP)} and \textit{credal sets} (e.g.,~\cite{w1991,a2014}) can be used to robustify our test towards deviations of its assumptions. Credal sets -- and generally imprecise probabilities -- form a consequent generalization of classical probability theory, which also accounts for partial probabilistic knowledge. Indeed, there are various reasons why the i.i.d.~assumption can be violated, ranging from unobserved heterogeneity to dependencies arising from data collection. The latter reason is particularly prevalent in surveys, where the survey mode (e.g., phone, web, in-person) often results in unequal, and even outcome-dependent, chances of the units to be sampled. Although methods exist to tackle this, such as reweighting schemes or random routing, most of them come with flaws of their own kind. For example, \cite{b2014, bauer2016biases} shows that random routing may be substantially biased, leading to informatively distorted selection probabilities, hence non i.i.d.~data. 

\subsection{The Robustified Testing Framework}

The rough idea of our robustification is to not analyze the test statistic based on $\hat{\pi}_X$ and $\hat{\pi}_Y$ alone, but use neighbourhood models or, more generally, \textit{credal sets} $\mathcal{M}_X \ni\hat{\pi}_X$ and $\mathcal{M}_Y \ni\hat{\pi}_Y$ of candidate probability measures instead. Credal sets -- introduced in~\cite{l1974} -- model partial probabilistic information by the set of all non-contradictory probabilities and have gained popularity in machine learning (e.g.,~\cite{corani2008learning, lienen2021credal, sh2021,Jansen2022,rjsa2023}, see also the corresponding literature referenced as related work in Section~\ref{sec:intro}).

The concrete idea behind our robustification is that we allow our samples to be (potentially) biased. We assume that these biased samples are similar to the true ones in the sense that the associated true empirical laws are contained in the credal sets $\mathcal{M}_X $ and $\mathcal{M}_Y $ around the biased empirical laws, respectively. We start by only assuming both $\mathcal{M}_X$ and $\mathcal{M}_Y$ to be (random) convex polyhedra with extreme points collected in the finite sets $\mathcal{E}(\mathcal{M}_X)$ and $\mathcal{E}(\mathcal{M}_Y)$. 

Now, we again want to test $H_0$ from~Eq.~(\ref{praghyp}), however, under the difficulty that the samples are biased. In the spirit of the concept of \textit{cautious data completion} (see, e.g.,~\cite[p.~181]{Augustin:Walter:Coolen:2014:itip:stat} or also~\cite{schollmeyer2019short} for the connections with stochastic dominance), one actually would adapt the resampling scheme discussed before by performing the test under all pairs of laws in the corresponding credal sets $\mathcal{M}_X $ and $\mathcal{M}_Y$. The null hypothesis $H_0$ from~Eq.~(\ref{praghyp}) would then be rejected whenever it is rejected for all such pairs.  Since this adapted resampling scheme is computationally cumbersome, we instead
look at the corresponding \textit{lower envelopes} $\underline{d}_{\mathbf{X},\mathbf{Y}}:\Omega \to \mathbb{R}$ and $\underline{d}^{\varepsilon}_{\mathbf{X},\mathbf{Y}}:\Omega \to \mathbb{R}$, respectively, given by
\begin{eqnarray*}
 \omega &\mapsto &  \inf_{(\pi_1 , \pi_2,u)\in \mathcal{D}}\sum_{z \in (\mathbf{X} \mathbf{Y})_{\omega}} u(z)\cdot(\pi_1(\{z\})-\pi_2(\{z\}))
\\
\omega &\mapsto&  \inf_{(\pi_1 , \pi_2,u)\in \mathcal{D}_{\varepsilon}}\sum_{z \in (\mathbf{X} \mathbf{Y})_{\omega}} u(z)\cdot(\pi_1(\{z\})-\pi_2(\{z\}))
\end{eqnarray*}
with $\mathcal{D}=\mathcal{M}^{\omega}_X\times \mathcal{M}^{\omega}_Y\times \mathcal{N}_{\mathcal{A}_{\omega}}$, $\mathcal{D}_{\varepsilon}=\mathcal{M}^{\omega}_X\times \mathcal{M}^{\omega}_Y\times \mathcal{N}^{\delta_{\varepsilon}(\omega)}_{\mathcal{A}_{\omega}}$ and $\mathcal{M}^{\omega}_X$ resp. $\mathcal{M}^{\omega}_Y$ the empirical credal sets given $\omega \in \Omega$. We compare these lower envelopes with the distribution (in the resamples) of the corresponding upper envelopes, $\overline{d}_{\mathbf{X},\mathbf{Y}}$  and $\overline{d}^{\varepsilon}_{\mathbf{X},\mathbf{Y}}$, that are obtained by replacing the part of $\inf$ concerning $\mathcal{M}^{\omega}_X\times \mathcal{M}^{\omega}_Y$ with the respective $\sup$ in the above definitions.
This gives a conservative yet valid statistical test.}

\subsection{Computation of $\mathbf{\underline{d}_{\mathbf{X},\mathbf{Y}}}$ and $\mathbf{\underline{d}^{\varepsilon}_{\mathbf{X},\mathbf{Y}}}$}
We now give an algorithm for the robustified test statistics.
\begin{prop}\label{its}
For $\mathbf{x}$ and $\mathbf{y}$ of form ~(\ref{sample1}) and~(\ref{sample2}), $\varepsilon \in [0,1]$, and $(\pi_1 , \pi_2) \in \mathcal{E}(\mathcal{M}^{\omega_0}_X) \times\mathcal{E}(\mathcal{M}^{\omega_0}_Y)$, consider the LP
\begin{equation}
    \sum_{\ell =1}^{s} v_{\ell}\cdot(\pi_1(\{z\})- \pi_2(\{z\})) \longrightarrow \min_{(v_1 , \dots , v_s)}
\end{equation}
with $(v_1 , \dots , v_s) \in C_{\varepsilon  \xi^*}(\mathbf{x},\mathbf{y})$ and $\xi^*$ the optimum of~(\ref{maxdel}).  Call opt$_{\varepsilon}(\mathbf{x},\mathbf{y},\pi_1, \pi_2)$ its optimum and $\underline{opt}_{\varepsilon}(\mathbf{x},\mathbf{y})$ the minimal optimum over $(\pi_1 , \pi_2) \in \mathcal{E}(\mathcal{M}^{\omega_0}_X) \times\mathcal{E}(\mathcal{M}^{\omega_0}_Y)$. Then:
\begin{itemize}
    \item[i)] $ \underline{opt}_{\varepsilon}(\mathbf{x},\mathbf{y})=\underline{d}^{\varepsilon}_{\mathbf{X},\mathbf{Y}}(\omega_0)$. 
    \item[ii)] There is in-sample GSD of $X$ over $Y$ for any $\pi$ with $\hat{\pi}^{\omega_0}_X \in \mathcal{M}^{\omega_0}_X$ and $\hat{\pi}^{\omega_0}_Y \in \mathcal{M}^{\omega_0}_Y$  if $\underline{opt}_0(\mathbf{x},\mathbf{y})\geq 0$.
\end{itemize}
\end{prop}

Proposition~\ref{its} requires to solve $ |\mathcal{E}(\mathcal{M}^{\omega_0}_X)| \cdot |\mathcal{E}(\mathcal{M}^{\omega_0}_Y)|$ linear programs. Depending on the concrete neighbourhood models, this is obviously limited: The number of programs is simply too large. A common strategy in such a case is to additionally assume 2-monotonicity of the considered credal sets, since this allows us -- at least for $R_1$ complete -- to give closed formulas for the upper and lower expectations. Unfortunately, this is not so simple in the case of a partially ordered $R_1$: since the representation via the Choquet integral (e.g.,~\cite{d1994}) depends on the order of elements of $A$, an optimum over all linear extensions of $R_1$ is needed to determine the most extreme Choquet integrals. In the worst case, this would lead to optimizing a non-convex function and thus hardly simplify the original problem (see~\cite{t2012}).

Another strategy is restricting to credal sets with  moderately many extreme points. We now consider one such possibility, namely the the \textit{$\gamma$- contamination model} (or \textit{linear-vacuous model}, see, e.g.,~\cite[p.~147]{w1991}). Here, we assume that for $\omega \in \Omega$, $\gamma \in [0,1]$, and $Z \in \{X,Y\}$ fixed, we have
\begin{equation} \label{lvm}
\mathcal{M}^{\omega}_Z =\Bigl\{ \pi: \pi \geq (1 -\gamma)\cdot \hat{\pi}^{\omega}_Z\Bigr\},
\end{equation}
where $\geq$ is understood event-wise. For 
$\gamma$-contamination models there are exactly as many extreme points as there are observed distinct data points, concretely given by
\begin{equation} \label{elvm}
    \mathcal{E}(\mathcal{M}^{\omega}_Z) = \Bigl\{ \gamma \delta_{z} + (1 - \gamma) \hat{\pi}^{\omega}_Z : \exists j \text{ s.t. } Z_j(\omega)=z \Bigr\},
\end{equation}
where $\delta_z$ denotes the Dirac-measure in $z$ (see again \citet[p.~147]{w1991}). Proposition~\ref{prop:test_statistic_ip} states that if the credal sets are both $\gamma$-contamination models, then a least favorable pair of extreme points can a priori be specified. The test statistics thus can be computed by solving one linear program.
\begin{prop}\label{prop:test_statistic_ip}
Consider again the situation of Proposition~\ref{its}, where additionally $\mathcal{M}^{\omega_0}_X$ and $\mathcal{M}^{\omega_0}_Y$ are of the form~(\ref{lvm}) with extreme points as in~(\ref{elvm}). It then holds:
$$\underline{opt}_{\varepsilon}(\mathbf{x},\mathbf{y})=opt_{\varepsilon}(\mathbf{x},\mathbf{y},\pi_*, \pi^*), \mbox{where}
$$
$$\pi_*=\gamma \delta_{a_*} + (1 - \gamma) \hat{\pi}^{\omega_0}_X~\text{ and }~\pi^*=\gamma \delta_{a^*} + (1 - \gamma) \hat{\pi}^{\omega_0}_Y.$$
\end{prop}
\section{Multidimensional spaces with differently scaled dimensions} \label{multi}
We now turn to a special case that is very common in applied research: multidimensional spaces whose dimensions may be of different scale of measurement.\footnote{For recent applications of such special preference systems to classifier comparison or multi-target decision making see~\cite{jnsa2022,ja2022} and~\cite{jsa2022}.} While traditional empirical research and policy support (e.g., \cite{EC:Eurostat:Comp:Indicators:2023}) summarizes such situations by  
indices/indicators that suffer eo ipso from ``the subjectivity of choices associated with them'' (\cite[p.~11]{UNECE:Ind.2019}), the embedding into the framework considered here allows a faithful representation of the entire underlying information. 

Concretely, we address $r \in \mathbb{N}$ dimensional spaces for which we assume --  w.l.o.g.~-- that the first $0 \leq z \leq r$ dimensions are of cardinal scale (implying that differences of elements may be interpreted as such), while the remaining ones are purely ordinal (implying differences to be meaningless apart from the sign). Specifically, we consider (bounded subsystems of) the preference system\footnote{One easily verifies that $R_1^*$ and $R_2^*$ are preorders. 
}
\begin{equation}\label{cps}
\textit{\textsf{pref}}(\mathbb{R}^r)=[ \mathbb{R}^r,R_1^*,R_2^*]   
\end{equation}
where
\begin{eqnarray*}
 R_1^*=\Bigl\{(x,y): x_j \geq y_j ~\forall j\leq r \Bigr\},~\text{and} ~~~~~~~~~~~~~~~~~~~~~~~~~~~~~~~~
\\
 R_2^*  =  \Biggl\{((x,y),(x',y')): \begin{array}{lr}
        x_j -y_j \geq x_j'-y_j' ~~\forall j\leq z\\
        x_j \geq x_j'\geq y_j' \geq  y_j ~\forall j> z
    \end{array}\Biggr\}.
\end{eqnarray*}
While $R_1^*$ can be interpreted as a simple component-wise dominance relation, $R_2^*$ deserves some more explanation: One pair of consequences is preferred to another one if it is ensured in the ordinal dimensions that the exchange associated with the first pair is not a deterioration to the exchange associated with the second pair and, in addition, there is component-wise dominance of the differences of the cardinal dimensions. The following proposition lists some important results for a more precise characterization of the GSD-relation on multidimensional structures.
\begin{prop} \label{somecharacterisitcs}
Let $\pi$ be a probability measure on $(\Omega,\mathcal{S}_1)$, and  $X=(\Delta_1 , \dots ,\Delta_r),Y=(\Lambda_1 , \dots ,\Lambda_r) \in \mathcal{F}_{(\textsf{pref}(\mathbb{R}^r),\pi)}$.  Then, the following holds:
\begin{itemize}
    \item[i)] $\textsf{pref}(\mathbb{R}^r)$ is consistent.
     \item[ii)] If $z=0$, then $R_{(\textsf{pref}(\mathbb{R}^r),\pi)}$ equals (first-order) stochastic dominance w.r.t.~$\pi$ and~$R_1^*$ (short: FSD$(R_1^*,\pi)$).
      \item[iii)] If $(X,Y) \in R_{(\textsf{pref}(\mathbb{R}^r),\pi)}$ and $\Delta_j, \Lambda_j \in \mathcal{L}^1(\Omega ,\mathcal{S}_1, \pi)$ for all $j=1, \dots, r$, then
      \begin{itemize}
          \item[I.] $\mathbb{E}_{\pi}(\Delta_j) \geq \mathbb{E}_{\pi}(\Lambda_j)$ for all $j=1, \dots , r$, and
          \item[II.] $(\Delta_j,\Lambda_j) \in$FSD$(\geq , \pi)$ for all $j=z+1, \dots , r$.
      \end{itemize}
      Additionally, if all components of $X$ are jointly independent and all components of $Y$ are jointly independent, properties I. and II. imply $(X,Y) \in R_{(\textsf{pref}(\mathbb{R}^r),\pi)}$.
\end{itemize}
\end{prop}
Part iii) of Proposition~\ref{somecharacterisitcs} is complete in the sense that the addition actually holds only under stochastic independence. 
\begin{rem}
The addition to iii) does not generally hold. A counterexample is
$z=1$, $r=2$, $\Omega=\{\omega_1,\ldots,\omega_4\}$, and $\pi$ the uniform distribution over $\Omega$. Then, for $\Delta_1(\omega) = 1,1,2,2$, $\Delta_2(\omega)=1,1,2,2$,  $\Lambda_1(\omega)=1,1,2,2$, and $\Lambda_2(\omega)=1,2,1,2$ for $\omega=\omega_1,\ldots, \omega_4$, it holds that $\mathbb{E}_{\pi}(\Delta_1)=\mathbb{E}_{\pi}(\Lambda_1)$. In fact, the first components are  equivalent with respect to first order stochastic dominance. The same holds for the second components. However, the whole vectors are incomparable with respect to first order stochastic dominance, since there is no corresponding mass transport from higher values to lower (or equal) values possible. Additionally, for 
 $u(x,y):=x\cdot y$, we have that $u \in\mathcal{U}_{ \textsf{pref}(\mathbb{R}^r)}$, $\mathbb{E}_{\pi}(u\circ\Delta)=10/4$, whereas $\mathbb{E}_{\pi}(u\circ\Lambda)=9/4$. Thus, $\Delta$ and $\Lambda$ can not be equivalent with respect to GSD.
\end{rem}
As an immediate consequence  of Proposition~\ref{somecharacterisitcs}, we have the following corollary for bounded subsystems of $\textsf{pref}(\mathbb{R}^r)$.
\begin{cor} \label{cor1}
If $\mathcal{C}=[C,R_1^c,R^c_2]$ is a bounded subsystem of $\textsf{pref}(\mathbb{R}^r)$ and $X,Y \in \mathcal{F}_{(\mathcal{C},\pi)}$, then $\mathcal{C}$ is $0$-consistent and ii) and iii) from Prop.~\ref{somecharacterisitcs} hold, if we replace $R_{(\textsf{pref}(\mathbb{R}^r),\pi)}$ by $R_{(\mathcal{C},\pi)}$, FSD$(R_1^*,\pi)$ by FSD$(R_1^c,\pi)$, and $(X,Y) \in R_{(\textsf{pref}(\mathbb{R}^r),\pi)}$ by $\forall u \in \mathcal{N}_{\mathcal{C}}: \mathbb{E}_{\pi}(u \circ X) \geq \mathbb{E}_{\pi}(u \circ Y)$.
\end{cor}
Finally, we give a characterization of the set of all representations of $\textsf{pref}(\mathbb{R}^r)$ if only one dimension is cardinal.
\begin{prop} \label{simplification}
Let $z=1$ and denote by $\mathcal{U}_{sep}$ the set of all $u: \mathbb{R}^r \to \mathbb{R}$ such that, for $(x_2 , \dots , x_r) \in \mathbb{R}^{r-1}$ fixed, the function $u(\cdot , x_2 , \dots , x_r)$ is strictly increasing and (affine) linear and such that, for  $x_1 \in \mathbb{R}$ fixed, the function $u(x_1, \cdot, \dots , \cdot)$ is strictly isotone w.r.t.~the the componentwise partial order on $\mathbb{R}^{r-1}$. Then $\mathcal{U}_{sep}=\mathcal{U}_{\textsf{pref}(\mathbb{R}^r)}$.
\end{prop}
\section{Applications} \label{application}
We now apply our framework on three examples: dermatological symptoms, credit approval data, and multidimensional poverty measurement. Results from the former two applications are presented in Supp.~D, while Section \ref{sec:poverty} discusses results from poverty analysis. Before that, some details on the concrete implementation are given.

\subsection{Implementation} \label{implementation}

To compute the test statistics for sample size $s$, we use a LP with constraints given by $C(x,y)$ (Section~\ref{sec:compu_test_statistic}). The computation of the test statistics and the maximum regularization strength $\xi^*$, see Proposition~\ref{prop:test_statitic_reg} and \ref{prop:xi_computation}, are LPs based on this same constraint matrix. The robustified statistics under $\gamma$-contamination are shifted versions of the original ones. Here, we utilize the linear connection between $\underline{d}^{\varepsilon}_{\mathbf{X},\mathbf{Y}}(\omega_0)$ and $ d^{\varepsilon}_{\mathbf{X},\mathbf{Y}}(\omega_0)$, $\overline{d}_I^{\varepsilon}$ and $d_I^{\varepsilon}$, respectively, for fixed $\epsilon$ (see Supp.~C).

Although one only needs to compute the constraint matrix once, the worst-case complexity of the computation is $\mathcal{O}(s^4)$. 
In the implementation, we focused on the case of two ordinal variables and only one numerical variable, using the preference system (\ref{cps}). We exploit the fact that sorting the data set allows some comparisons to be skipped immediately by considering only the ordinal components. In particular, if the ordinal variables have a small number of categories compared to the sample size $s$, this can lead to a large proportion of comparisons being skipped. In the most cases, this reduces the computational cost of computing the constraint matrix compared to a naive implementation. Of course, in the worst case, the computation time cannot be drastically reduced in this way. For further details on the implementation, see Supp.~B.

\subsection{Example: Poverty analysis} 
\label{sec:poverty}
At least since the capability approach by \cite{sen1985commodities}, there is mostly consensus that poverty has more facets than income or wealth. It is perceived as multidimensional concept, involving variables that are often ordinally scaled, e.g., level of education. One common task in poverty analysis is to compare subgroups like men and women. Stochastic dominance is a popular way of comparing such subpopulations, see e.g. \cite{garcia2019review}. Excitingly, our approach allows us to extend this to multidimensional poverty measurement with any kind (of scales) of dimensions. 

\begin{figure}[h!]
    \centering
    \includegraphics[width=(\columnwidth-1.4cm)]{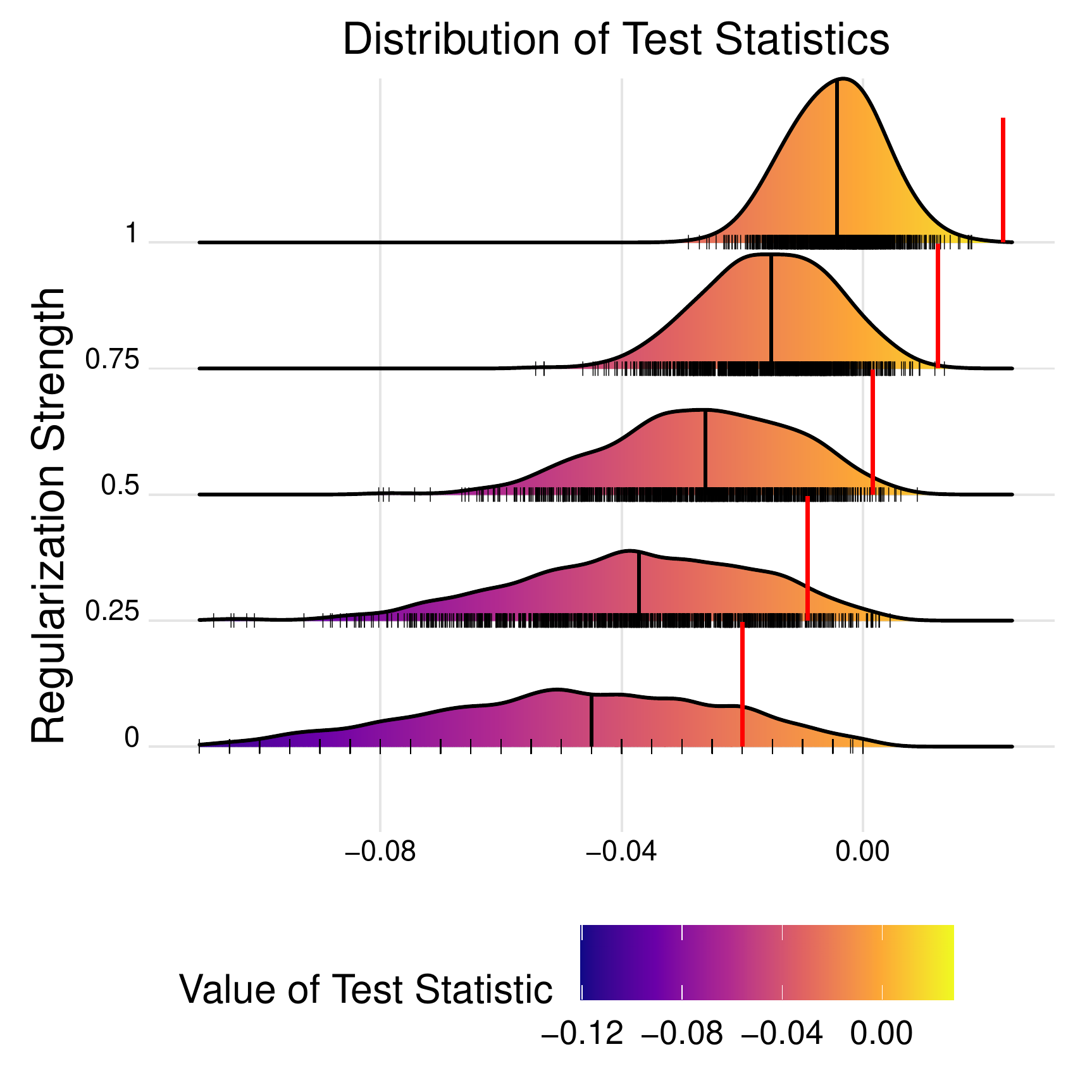}
    \caption{Distributions of ${d}^{\varepsilon}_{I}$ with $\varepsilon \in \{0,0.25,0.5,0.75,1\}$ 
     obtained from $N = 1000$ resamples of ALLBUS data. Black stripes show exact positions of ${d}^{\varepsilon}_{I}$ values. Vertical black line marks median. Red line shows value of the respective observed test statistics ${d}^{\varepsilon}_{\mathbf X, \mathbf Y}(\omega) $.  }
    \label{fig:res-distr}
\end{figure}

In the following, we will use data from the German General Social Survey (ALLBUS) \cite{allbus-2014} that accounts for three dimensions of poverty: income (numeric), health (ordinal, 6 levels) and education (ordinal, 8 levels), see also \cite{breyer2015skala}. We are using the 2014 edition and focus on a subsample with $n = m = 100$ men and women each. We are interested in the hypothesis that women are dominated by men with respect to GSD -- differently put, that women are poorer than men regarding any compatible utility representation of income, health and education. 

As discussed in Section~\ref{testdom}, we test the hypotheses (\ref{praghyp}), where $X$ resp. $Y$ correspond to the subpopulation of men resp. women. We deploy our test with varying regularization strength~$\varepsilon$. Figure \ref{fig:res-distr} displays the distribution of the test statistics obtained trough $N = 1000$ resamples (cf.~Section \ref{sec:compu_test_statistic}).
It becomes evident that our proposed regularization serves its purpose: As $\varepsilon$ increases, the distribution of tests statistics becomes both more centered and closer to zero. Moreover, we reject for higher shares of the test statistics, see the position of ${d}^{\varepsilon}_{\mathbf X, \mathbf Y}(\omega)$ (red line) compared to ${d}^{\varepsilon}_{I}$ (black stripes). For~$\varepsilon \in \{0.5,0.75,1\}$ we reject for the common significance level of $\alpha \approx 0.05$. 

\begin{figure}[h!]
    \centering
    \includegraphics[scale=0.25]{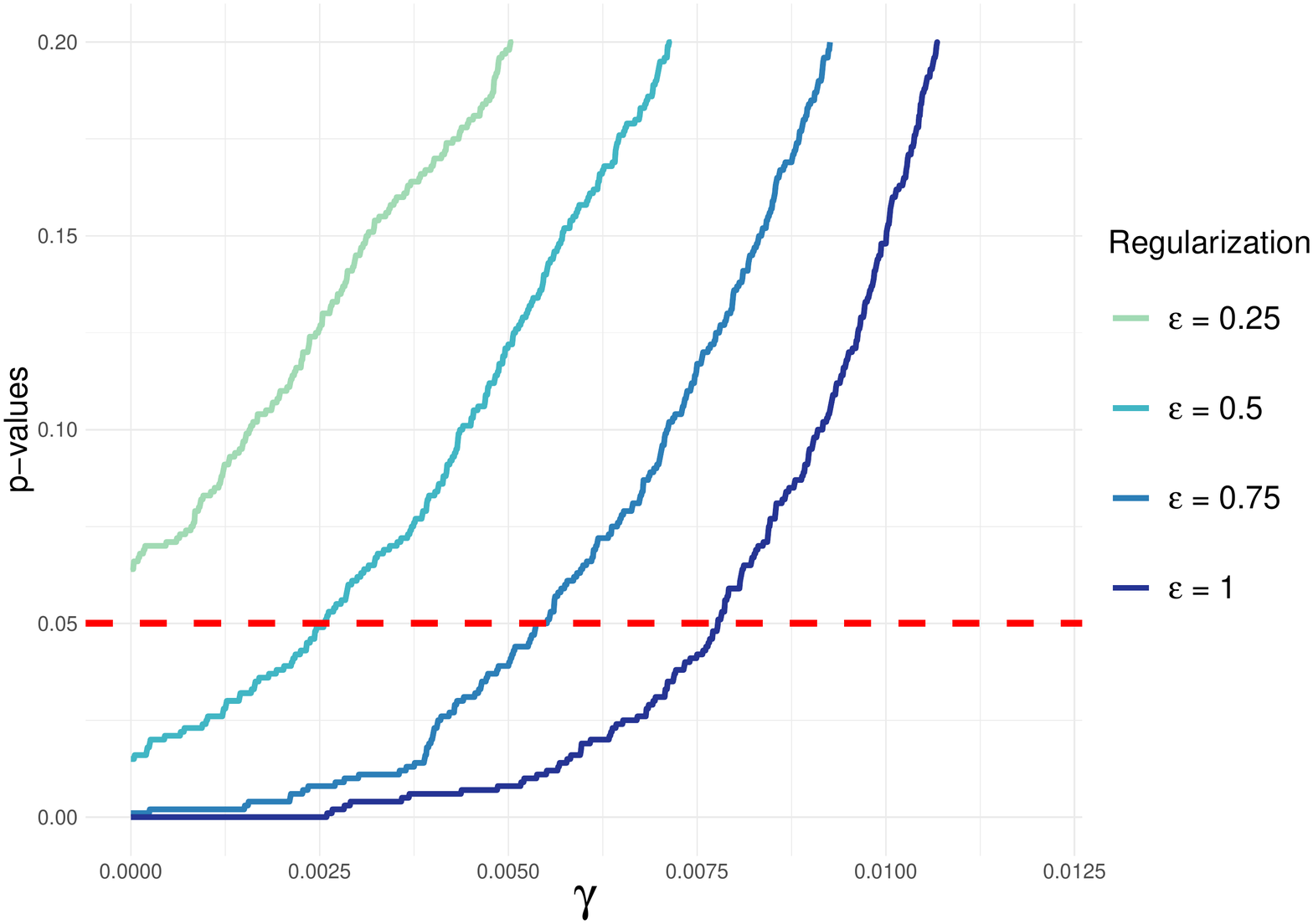}
    \caption{P-values as function of the contamination~$\gamma$ (see Supp.~C) for tests with different regularization strength~$\varepsilon$. Dotted red line marks significance level $\alpha = 0.05$. 
    }
    \label{fig:res-share-rej}
\end{figure}

As touched upon in Section~\ref{implementation}, the robustified versions of the test statistic under the linear-vacuous model are shifted versions of the regular test statistics, i.e., they do not have to be computed explicitly. Exploiting this fact, we visualize the share of regularized test statistics for which we do not reject the null hypothesis (black stripes right of red line in Figure~\ref{fig:res-distr}), depending on the contamination parameter $\gamma$ of the underlying linear-vacuous model, see Figure~\ref{fig:res-share-rej} (and Supp.~C for details on computing the shares). It should be mentioned that these shares correspond to p-values telling at which significance levels $\alpha$ the test would be marginally rejected.   
Generally, it becomes apparent that even for small values of~$\gamma$ the test statistics can be severely corrupted. If we allow more than $1\%$ ($\gamma > 0.01$) of the data ($2$ observations) to be redistributed in any manner, the shares of rejections drop drastically. Therefore, ignoring an (even very tiny) contamination $\gamma$ of the underlying distributions leads to a seriously inflated type~I error. 
 Remarkably, our regularization hedges against this to some extent:  
Given a significance level $\alpha=0.05$, the fully regularized version (i.e., $\varepsilon=1$) of our robustified test (cf., Section~\ref{robtest})  comes to the same decision for $\gamma$ up to $0.075$. As explained in Section~\ref{testdom}, rejecting $H_0$ does not necessarily mean that women are dominated by men; they could also be incomparable. However, our tests with reversed variables give no evidence of incomparability: all their observed p-values are above $0.95$.

\textbf{Further Applications:}
We also analyzed a dermatology data set that contains variables on symptoms of the eryhemato-squamous disease, see \cite{demiroz1998learning} accessed via \cite{Dua:2019}, as well as the German credit data set that consists of variables on credit applicants, see \cite{Dua:2019}. In case of the credit data, we reject the hypothesis that high-risk applicants are dominated by low-risk applicants w.r.t. GSD for a common significance level of $\alpha \approx 0.05$. In the first application we are interested in the hypothesis that patients without a family history of the disease are dominated by patients without a family history with respect to GSD. We reject again for $\alpha \approx 0.05$. However, the p-values are much higher than in the other two applications. For detailed results as well as more information on the data sets, we refer to the supplement.
\section{Concluding Remarks} \label{conrem}
\textbf{Summary:} We have further explored a generalized stochastic dominance (GSD) order among random variables with locally varying scale of measurement. We focused on four aspects: First, the investigation of (regularized) statistical tests for GSD when only samples of the variables are available. Second, robustifications of these tests w.r.t.~their underlying assumptions using ideas from imprecise probabilities. Third, a detailed investigation of our ordering for preference systems arising from multidimensional structures with differently scaled dimensions. Finally, applications to examples from poverty measurement, finance, and medicine. 

\textbf{Limitations and future research:} Two particular limitations offer promising opportunities for future research.

 \textit{Extending robust testing to belief function:} In Section~\ref{robtest}, we have focused -- for computational complexity -- to linear-vacuous models. However, the idea of identifying least favorable extreme points seems to generalize to any credals sets induced by belief functions in the sense of~\cite{s1976}.

\textit{Improving computational complexity:} The LPs for checking in-sample GSD become computer intensive for larger amounts of data. Although complexity reduces for the special case of preference systems discussed in Section~\ref{multi} (cf.~Section~\ref{implementation}), Proposition~\ref{simplification} suggests that a further drastic reduction can be expected for only one cardinal dimension. 

\begin{acknowledgements}
We thank all reviewers for constructive comments that helped to improve the presentation of the paper. JR and TA gratefully acknowledge support by the Federal Statistical Office of Germany within the project "Machine Learning in Official Statistics". HB, JR, and GS gratefully acknowledge the financial and general support of the LMU Mentoring program. HB sincerely thanks Evangelisches Studienwerk e.V. for funding and supporting her PhD project.
\end{acknowledgements}

\bibliography{jansen_234}

\end{document}